\documentclass[journal]{IEEEtran}
\usepackage{amsmath,amsfonts}
\usepackage{algorithmic}
\usepackage{algorithm}
\usepackage{array}
\usepackage[caption=false,font=footnotesize,labelfont=rm,textfont=rm]{subfig}
\usepackage{textcomp}
\usepackage{stfloats}
\usepackage{url}
\usepackage{verbatim}
\usepackage{graphicx}
\usepackage{subfig}
\usepackage{graphicx} 
\usepackage{cite}
\usepackage[colorlinks=true, linkcolor=blue, citecolor=blue, urlcolor=red]{hyperref}
\usepackage{multirow}
\usepackage{booktabs}
\usepackage{makecell}
\usepackage{xcolor}

\begin{document}

\title{Automated Neural Architecture Design for \\ Industrial Defect Detection}

\author{Yuxi Liu, Yunfeng Ma, Yi Tang, Min Liu, Shuai Jiang, and Yaonan Wang
\thanks{This work was supported in part by the National Natural Science Foundation of China under Grant 62221002 and 62425305, in part by the Natural Science Foundation of Hunan Province under Grant 2024JJ3013, in part by the Science and Technology Innovation Program of Hunan Province under Grant 2023RC1048. (Yuxi Liu and Yunfeng Ma contributed equally to this work. Corresponding author: Min Liu).}

\thanks{The authors are with the School of Artificial Intelligence and Robotics at Hunan University and the National Engineering Research Center for Robot Visual Perception and Control Technology, Changsha, 410082, Hunan, China. E-mails: \{yuxi\_liu, ismyf, tyhnu, liu\_min, svyj, yaonan\}@hnu.edu.cn}}

%
\markboth{}%
{Shell \MakeLowercase{\textit{et al.}}: A Sample Article Using IEEEtran.cls for IEEE Journals}

\maketitle
\begin{abstract}
Industrial surface defect detection (SDD) is critical for ensuring product quality and manufacturing reliability. Due to the diverse shapes and sizes of surface defects, SDD faces two main challenges: intraclass difference and interclass similarity. Existing methods primarily utilize manually designed models, which require extensive trial and error and often struggle to address both challenges effectively. To overcome this, we propose AutoNAD, an automated neural architecture design framework for SDD that jointly searches over convolutions, transformers, and multi-layer perceptrons. This hybrid design enables the model to capture both fine-grained local variations and long-range semantic context, addressing the two key challenges while reducing the cost of manual network design. To support efficient training of such a diverse search space, AutoNAD introduces a cross weight sharing strategy, which accelerates supernet convergence and improves subnet performance. Additionally, a searchable multi-level feature aggregation module (MFAM) is integrated to enhance multi-scale feature learning. Beyond detection accuracy, runtime efficiency is essential for industrial deployment. To this end, AutoNAD incorporates a latency-aware prior to guide the selection of efficient architectures. The effectiveness of AutoNAD is validated on three industrial defect datasets and further applied within a defect imaging and detection platform. Code is available at https://github.com/Yuxi104/AutoNAD.
\end{abstract}


\section{Introduction}
\IEEEPARstart{S}{urface} defect detection (SDD) is a critical task in industrial automation, ranging from quality control in manufacturing~\cite{tmech-autoencoder, tmech-multi-view, tmech-robust} to maintenance in infrastructure\cite{tase-RDNet-KD}. To ensure reliable inspection, many practical systems adopt pixel-level detection strategies~\cite{tip-wavelet}, enabling accurate localization and contour extraction of defects. However, achieving robust detection remains difficult due to two core challenges: intraclass difference and interclass similarity~\cite{PGA,tim-boundary}.

\begin{figure} [t]
    \centering
    \subfloat[Intraclass difference\label{challenge_a}]{
        \includegraphics[width=0.47\textwidth]{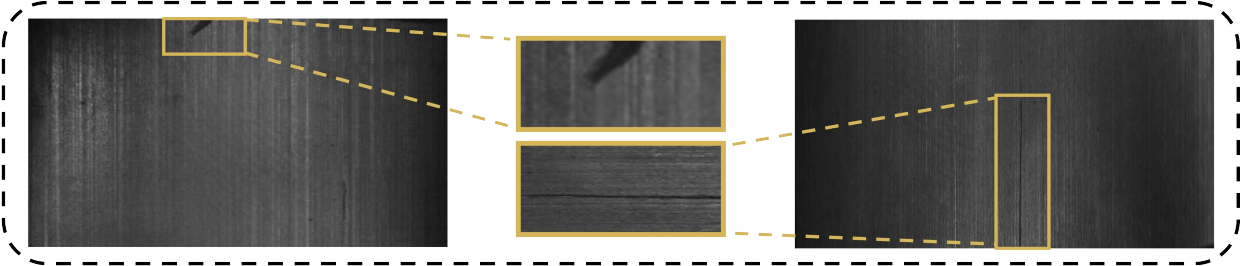}}
        \vspace{-1mm}
    \\
    \subfloat[Interclass similarity\label{challenge_b}]{
        \includegraphics[width=0.47\textwidth]{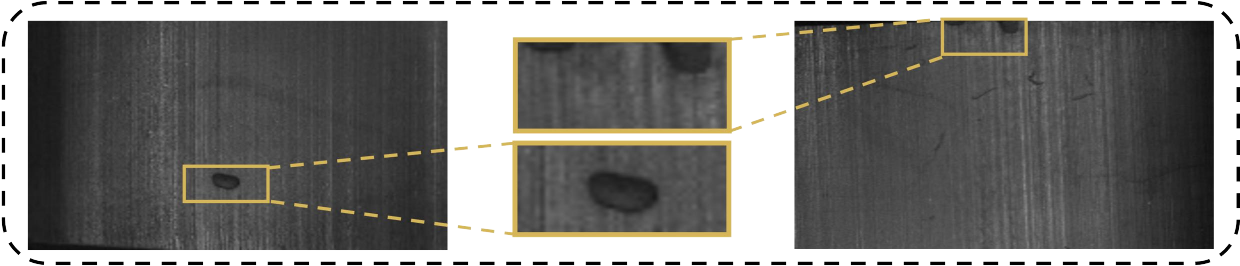} }
    \caption{Challenges of surface defect detection. (a) Intraclass difference. The shapes of crack defects vary significantly. (b) Interclass similarity. Blowhole and break defects exhibit considerable similarities in pattern.}
    \vspace{-0.6cm}
\end{figure}

The first challenge, intraclass difference, arises from diverse appearances (shape, size, and texture) within the same defect type, which can hinder consistent feature representation. The second challenge, interclass similarity, occurs when different categories exhibit similar patterns, often leading to ambiguous classifications. As shown in Fig.~\ref{challenge_a} and Fig.~\ref{challenge_b}, these challenges are particularly pronounced in real-world industrial environments, where defect appearances can vary across production batches, and visual similarities between defect types can mislead both human inspectors and automated detection systems.

Recent advances in deep learning have introduced CNNs, transformers, and MLPs into SDD pipelines \cite{tip-wavelet,tim-boundary,tim-global_local,tim-Context-Aware, tim-mlp}, each contributing unique capabilities. Convolutions extract local texture details~\cite{iccv-cvt} critical for modeling intraclass differences. Transformers capture global contextual relations~\cite{vits}, which helps disambiguate visually similar defect categories. MLPs, with their high-capacity abstraction, bridge global and local semantics. However, these architectures each have inherent limitations. CNNs often struggle with long-range dependencies, transformers may overlook subtle local variations, and MLP-based models are prone to overfitting on limited industrial data. These limitations hinder single-operator architectures from fully addressing the challenges in SDD.

A straightforward solution is to design a hybrid architecture that leverages the complementary strengths of the three types of operators, effectively addressing both challenges simultaneously. Nevertheless, manually designing such hybrid architectures is not trivial. Selecting and configuring the optimal combination of operators requires extensive domain knowledge and time-consuming trial-and-error. As a result, most existing SDD pipelines still adopt fixed, manually designed architectures that can only partially address the above challenges and often fail to fully exploit the potential of hybrid designs.

\begin{figure}[t]
    \centering
    \includegraphics[width=0.489\textwidth]{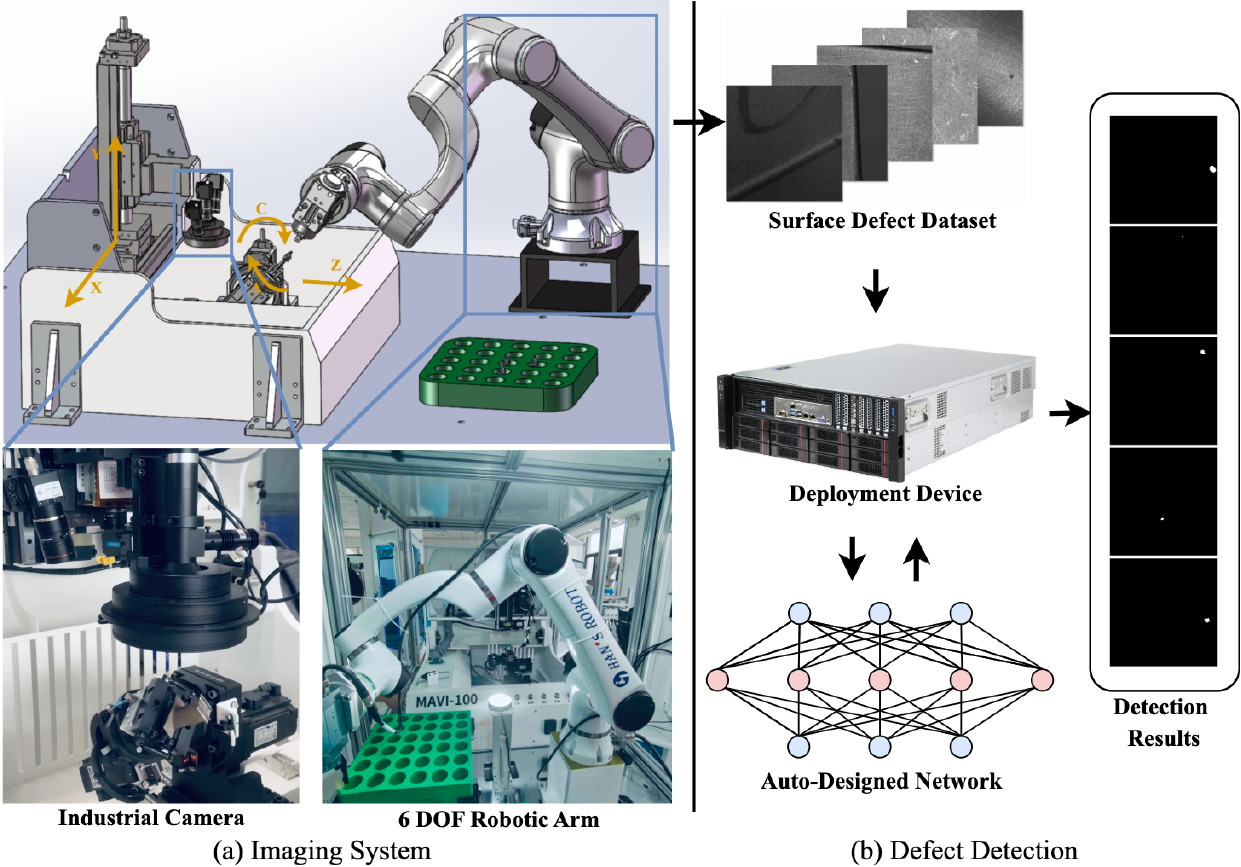} 
    \caption{The automated defect imaging and detection platform. It consists of two main parts: imaging and detection.}
    \label{platform}
    \vspace{-0.6cm}
\end{figure}

This labor-intensive and highly customized process fundamentally conflicts with the core objective of artificial intelligence: building systems that are automated, adaptive, and scalable. Why should ``intelligent" systems still depend heavily on manual effort to design suitable architectures? Why not develop an automated solution capable of designing models based on the characteristics and constraints of each industrial application?

To answer these questions, we propose AutoNAD (Automated Neural Architecture Design), a system for industrial defect detection based on neural architecture search (NAS). AutoNAD explores hybrid network designs combining convolution, transformer, and MLP operators. By constructing a hybrid supernet, it unifies diverse operator types within a comprehensive search space and enables adaptive composition of feature extractors tailored to specific dataset characteristics and deployment requirements.

To make the large-scale hybrid search tractable and efficient, we introduce a cross weight sharing strategy that integrates different operator types under a unified search space and synchronizes weight updates across the supernet. Moreover, only the largest operator block's weight is stored in each layer, while smaller blocks inherit weights from it. This design enables efficient weight sharing not only within the same operator type, but also across different types, thus accelerating convergence and improving the performance of subnets.

In addition, this work explicitly considers efficiency to better support practical deployment in industrial scenarios. Specifically, AutoNAD incorporates a latency-aware prior during the architecture search process. This prior is constructed from runtime statistics collected during supernet training and guides the search algorithm to favor architectures that balance detection accuracy with computational latency. Without introducing an latency predictor, this lightweight prior serves as an effective proxy for real-world resource constraints. Furthermore, prior studies~\cite{tim-global_local,tim-Deep-Feature} have shown that multi-level features are essential for SDD, where low-level features capture fine textures and high-level features offer semantic abstraction. To automatically adapt feature fusion to different defect types and image scales, we develop a searchable multi-level feature aggregation module (MFAM). Structured as a directed acyclic graph (DAG), MFAM dynamically selects optimal fusion paths across levels through the NAS process.

To further demonstrate its practical viability, AutoNAD has been integrated into a real-world automated imaging and detection platform (Fig.\ref{platform}), enabling end-to-end architecture design and deployment under actual production constraints. The detailed system integration is presented in Sec.\ref{application}.

Our contributions are summarized as follows:
\begin{itemize} 
    \item{We propose AutoNAD, an automated neural architecture design framework for surface defect detection. By leveraging hybrid architecture search and a latency-aware prior, AutoNAD enables adaptive network design tailored to diverse datasets and deployment scenarios, effectively addressing the challenges of intraclass difference and interclass similarity.} 
    \item{A unified search space is designed for different types of operators, where convolution, transformer, and MLP are formulated in the same format. Based on this, a cross weight sharing strategy is introduced to enable efficient weight sharing within and across operator types, which accelerates supernet convergence and improves subnet performance.} 
    \item{A searchable multi-level feature aggregation module is developed, with the feature refinement and aggregation process structured as a directed acyclic graph. This module enhances multi-scale feature learning by dynamically selecting optimal fusion paths.}
    \item{The proposed method is validated on three public defect datasets and further demonstrated through deployment in a real-world automated detection system, achieving high detection accuracy and efficiency.}
\end{itemize}

\section{Related Work}
\subsection{Pixel-level Surface Defect Detection}
Pixel-level surface defect detection (SDD) focuses on precisely identifying the contours and boundaries of defects. Existing approaches mainly rely on single-operator models (CNNs, transformers or MLPs), each focusing on different aspects of feature learning. CNN-based methods are well-suited for capturing fine-grained local patterns. For example, Wang et al.~\cite{tii-RERN} introduced RERN to enhance edge localization in solar cell defect images. CAWANet~\cite{tim-Context-Aware} employs adaptive weighted convolutions to improve detail preservation and reduce noise. Transformer-based approaches, leveraging self-attention mechanisms, excel in modeling long-range dependencies. CrackFormer~\cite{iccv-Crackformer} integrates scaling attention to suppress non-semantic background while enhancing defect semantics. PST~\cite{tip-wavelet} employs a two-stage transformer pipeline to achieve high detection accuracy through wavelet-based enhancement. Yuan et al.~\cite{tim-mlp} utilized MLP-Mixer blocks~\cite{nips-mlp-mixer} to capture contextual relationships beyond convolutional locality. While each operator has demonstrated unique advantages, existing SDD models are largely confined to single-operator paradigms. The application of hybrid architectures remains underexplored, particularly in the context of automated architecture design.

\subsection{Neural Architecture Search}
Neural architecture search (NAS) aims to automate the design of deep learning models, reducing reliance on manual tuning. Early NAS approaches typically rely on reinforcement learning~\cite{rl-1} or evolutionary algorithms~\cite{ea-1-icml}, which often incur high computational costs. To improve efficiency, one-shot NAS~\cite{one-shot-smash} have emerged. It builds an over-parameterized supernet where subnets share weights, significantly reducing training time. However, subnets often perform considerably below their true potential when compared to being trained from scratch, which affects the evaluation of the optimal network. AutoFormer~\cite{iccv-autoformer} solves this problem by entangling weights across transformer blocks, while BigNAS~\cite{eccv-bignas} stabilizes CNN supernets through the sandwich rule and in-place distillation. Despite their effectiveness, both methods remain architecture-specific, focusing exclusively on transformers or convolutions. One-shot NAS methods for segmentation, such as Auto-DeepLab~\cite{cvpr-Auto-DeepLab} and FasterSeg~\cite{fasterseg}, also rely on tailored CNN-based search spaces or manually designed multi-resolution schemes. However, industrial surface defect detection requires both fine-grained local detail and global contextual modeling, which single-operator search frameworks cannot fully capture. To address this gap, AutoNAD generalizes weight sharing to heterogeneous operators by unifying convolution, transformer, and MLP into a two-step operator form with cross-operator inheritance. In addition, AutoNAD integrates a predictor-free latency prior derived from runtime statistics and jointly searches the backbone together with a searchable MFAM, enabling improved subnet performance, faster convergence, and efficient deployment in industrial SDD scenarios.


\begin{figure*}[t]
    \centering
    \includegraphics[width=1.0\textwidth]{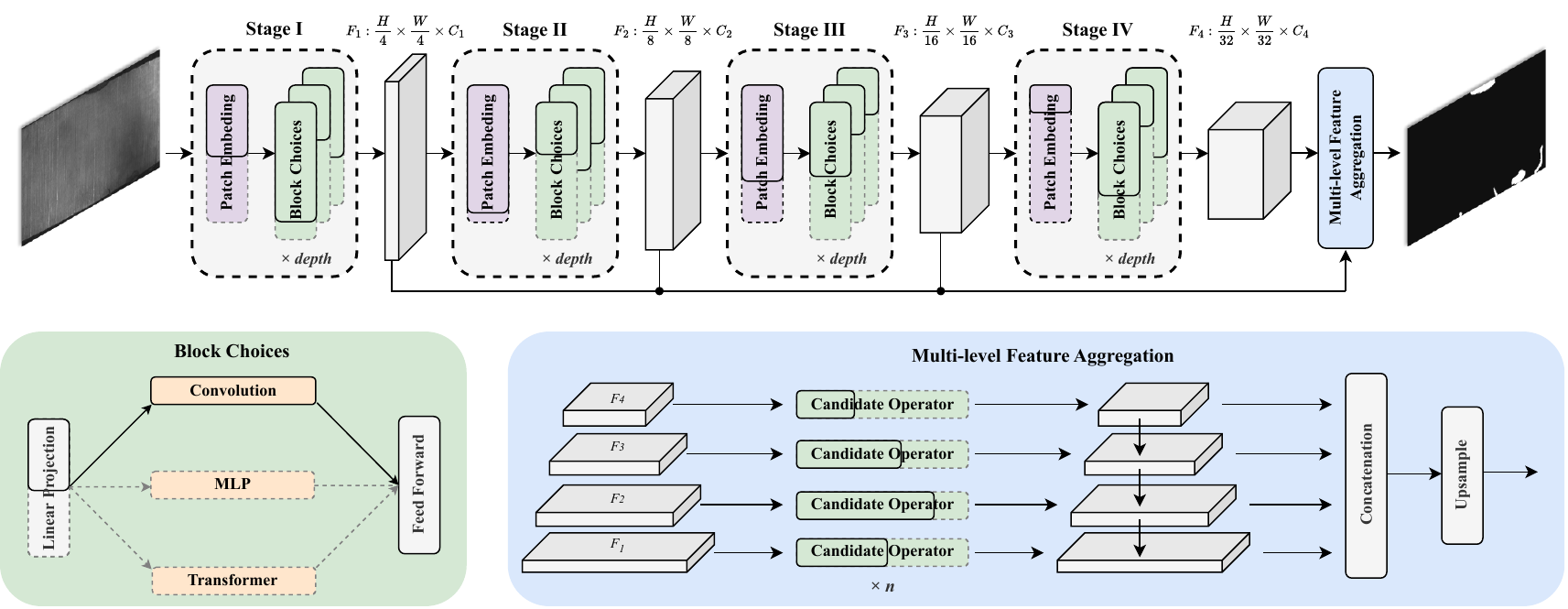} 
    \caption{The supernet architecture of the AutoNAD. The whole architecture is divided into two main parts: backbone and multi-level feature aggregation module. Each block's channel dimension is dynamically determined as part of the search process. Specifically, in the visualization, each block is depicted as a horizontal bar, with its total length representing the maximum channel width. The solid portion corresponds to the selected active channels, while the dashed portion indicates the unselected ones. Moreover, the depth of each block is also dynamic. Refer to Table~\ref{search space} and Sec.~\ref{mlfg search space} for more details about the search space.}
    \label{network}
    \vspace{-0.4cm}
\end{figure*}

\section{Preliminaries}
One-shot NAS\cite{single-one-shot,iccv-autoformer} separates the optimization of architecture from the training of parameters, treating them as two distinct stages: (1) Optimizes the parameters of all candidate subnets in the supernet through weight sharing. (2) Leverages common search algorithms, such as evolutionary algorithms, to identify the optimal subnet. Let $\mathcal{A}$ denote the search space, which is encoded into an over-parameterized supernet $\mathcal{N}(\mathcal{A}, W)$, where $W$ denotes the set of shared weights across all candidate subnets. The training objective for the first stage can then be formulated as:
\begin{equation}
W_{\mathcal{A}} = \operatorname*{arg\,min}_{W} \mathcal{L}_{\text{train}}(\mathcal{N}(\mathcal{A}, W)),
\end{equation}
where $\mathcal{L}_{\text{train}}$ is the loss function on the training set. Once the supernet has converged, the optimal subnet $\alpha^{*}$ can be obtained in the stage two:
\begin{equation}
    \alpha^{*} = \operatorname*{arg\,max}_{\alpha\in\mathcal{A}}\text{Acc}_\text{val}(\mathcal{N}(\alpha, w_{\alpha})),
\end{equation}
with $w_{\alpha}$ representing the weights inherited from $W_{\mathcal{A}}$ and $\text{Acc}_\text{val}$ denotes the performance evaluated on the validation set.

\section{Method}
The supernet architecture of AutoNAD is shown in Fig.~\ref{network}. It consists of two main components: a backbone and a multi-level feature aggregation module (MFAM). The backbone is divided into four stages based on downsampling ratios. Each stage includes configurable patch embedding choices input channels) and supports three types of blocks: convolution, transformer, and MLP. The MFAM conducts cell-level search over a predefined set of candidate operators commonly used in pixel-level defect detection, such as depthwise separable and dilated convolutions. To improve training efficiency and deployment performance, a latency-aware prior is integrated into the architecture search, using runtime statistics collected during supernet training to guide the selection of efficient subnets. In the following subsections, we introduce the unified search space and cross weight sharing strategy, followed by the design of MFAM and the latency-aware search pipeline.

\subsection{Unified Search Space for Convolution, Transformer and MLP}
To place three different types of operators into a unified weight sharing search space, the key is to convert them into the same format. We divide each operator into two main steps based on its own computation manner. The first step involves a unified linear projection process, and the second step will perform different paradigm operations. For transformer, it involves the computation of mult-head self-attention (MSA). For convoluiton, the projected feature maps are shifted and aggregated together. For MLP, we adopt the Spatial-Shift from S$^{2}$-MLPV2\cite{s2mlpv2} (a pure MLP architecture), which can achieve the communication between patches. Below, we introduce the unified formulation for each of them. 
\subsubsection{Self-Attention}
\label{Self-Attention}
Given a self-attention module with $\mathit{H}$ heads. Let $\mathit{X}\in\mathcal{R}^{c_{in}\times h\times w}$, $\mathit{Y}\in\mathcal{R}^{c_{out}\times h\times w}$ be the input and output feature maps, where $\mathit{h}$ and $\mathit{w}$ represent height and width of the feature map. $\mathit{y}_{i,j}$ and $\mathit{x}_{i,j}$ are the values of the feature maps $\mathit{Y}$ and $\mathit{X}$ at position $\mathit{(i,j)}$. The standard self-attention computation process can be formulated as:
\begin{equation}
\label{mhsa-01}
y_{i,j}=\mathrm{Concat}\left(x_{i,j}\mathrm{head}^{(1)},\cdots,x_{i,j}\mathrm{head}^{(H)}\right)W^{O},
\end{equation}
where
\begin{equation}
\label{mhsa-02}
x_{i,j}\mathrm{head}^{(l)} =\mathrm{Attention}\left(W_{q}^{(l)}x_{i,j},W_{k}^{(l)}x_{i,j},W_{v}^{(l)}x_{i,j}\right),
\end{equation}
\begin{equation}
\label{mhsa-03}
\mathrm{Attention}(Q,K,V)=\mathrm{softmax}\left(\frac{QK^{T}}{\sqrt{d_K}}\right).
\end{equation}
$W_{(l)}^{q}$, $W_{(l)}^{k}$, $W_{(l)}^{v}$ and $W^{O}$ are linear projection parameters. Based on above equations, self-attention module can be decomposed into two steps:
\begin{align}
\label{mhsa-04}
&\textbf{Step I}:{q}_{i,j}^{(l)}=W_{q}^{(l)}x_{i,j}, {k}_{i,j}^{(l)}=W_{k}^{(l)}x_{i,j},{v}_{i,j}^{(l)}=W_{v}^{(l)}x_{i,j},\\
\label{mhsa-05}
&\textbf{Step II}:{y}_{i,j}= \underset{l=1}{\overset{H}{\|}}\mathrm{Attention}\left({q}_{i,j}^{(l)},{k}_{i,j}^{(l)},{v}_{i,j}^{(l)}\right)W^{O}.
\end{align}
where ${\|}$ is the concatenation of $H$ attention heads. 
\subsubsection{Convolution}
Given a standard convolution, it can be described as $\mathit{K}\in\mathcal{R}^{c_{out}{\times c}_{in}\times k\times k}$, where $\mathit{k}$ denotes the kernel size and $\mathit{c}_{out}$, $\mathit{c}_{in}$ are the number of output and input channels. Let $\mathit{X}\in\mathcal{R}^{c_{in}\times h\times w}$, $\mathit{Y}\in\mathcal{R}^{c_{out}\times h\times w}$ be the input and output feature maps, where $\mathit{h}$ and $\mathit{w}$ represent width and height of the feature map. The standard convolution computation process can be formulated as:
\begin{equation}
\label{cnn-01}
y_{i,j}=\sum_{a=0}^{k-1}\sum_{b=0}^{k-1}{K_{(a,b)}x_{i+a,j+b}},
\end{equation}
where $\mathit{y}_{i,j}$ and $\mathit{x}_{i,j}$ are the values of the feature maps $\mathit{Y}$ and $\mathit{X}$ at position $\mathit{(i,j)}$. $\mathit{a,b}$ represents the indices of the kernel position and $\mathit{K_{(a,b)}}$ denotes the kernel weights at $\mathit{(a,b)}$.

To facilitate the subsequent derivation of formulas, the Eq.~\eqref{cnn-01} can be rewrote as the summation of the shifted windows, here we first defined the $\mathrm{shift}$ operation:
\begin{equation}
\label{cnn-02}
\mathrm{shift}(x_{i,j},\Delta m,\Delta n)={x}_{i+\Delta m,j+\Delta n},
\end{equation}
where $\Delta m$ and $\Delta n$ are the shift operations associated with the horizontal and vertical displacements. Then, the summation of the shifted windows can be described as:
\begin{equation}
\label{cnn-03}
y_{i,j}= \sum_{a,b}{\mathrm{shift}\left(K_{(a,b)}x_{i,j},a,b\right)},
\end{equation}
Based on the Eq.~\eqref{cnn-03}, convolution can be divided into two steps:
\begin{align}
\label{cnn-04}
&\widetilde{y}_{i,j}^{(a,b)}=K_{(a,b)}x_{i,j}, \\
\label{cnn-05}
&y_{i,j}= \sum_{a,b}{\mathrm{shift}\left(\widetilde{y}_{i,j}^{(a,b)},a,b\right)}.
\end{align}

In the first step, we use the given kernel to linearly project the feature maps. Inspired by ACmix\cite{cvpr-on-the-integra}, this operation can be easily replaced by standard $ 1\times1 $ convolutions. To be consistent with the format of self-attention, we using three $ 1\times1 $ convolution to substitute step 1, with the resulting three outputs corresponding to $q$, $k$, and $v$ in Eq.~\eqref{mhsa-04}:
\begin{flalign}
\label{cnn-06}
\begin{split}
    \textbf{Step I}:\widetilde{y}_{i,j}^{(1)}&=K_{(a,b)}^{(1)}x_{i,j},\widetilde{y}_{i,j}^{(2)}=K_{(a,b)}^{(2)}x_{i,j}, \\
    \widetilde{y}_{i,j}^{(3)}&=K_{(a,b)}^{(3)}x_{i,j}, 
\end{split}&
\end{flalign}

\begin{flalign}
\label{cnn-07}
\textbf{Step II}:&\widetilde{y}_{i,j}^{(a,b)}=\mathrm{Concat}\left(\widetilde{y}_{i,j}^{(1)},\widetilde{y}_{i,j}^{(2)},\widetilde{y}_{i,j}^{(3)}\right)W^{P},&&\\
&y_{i,j}= \sum_{a,b}{\mathrm{shift}\left(\widetilde{y}_{i,j}^{(a,b)},a,b\right)}.
\end{flalign}
where $W^{P}$ is linear projection parameter.
\subsubsection{Spatial-Shift}
Let $\mathit{X}\in\mathcal{R}^{c_{in}\times h\times w}$ and $\mathit{Y}\in\mathcal{R}^{c_{out}\times h\times w}$ denote the input feature and output feature maps. We first expand the channels of $\mathit{X}\in\mathcal{R}^{c_{in}\times h\times w}$ from $c_{in}$ to $3c_{in}$ by an standard MLP operation: 
\begin{equation}
\label{mlp-01}
\widetilde{X} = \mathrm{MLP}(X).
\end{equation}
where $\widetilde{X}\in\mathcal{R}^{3c_{in}\times h\times w}$. Second, $\widetilde{X}$ is equally split along the channel dimension into three parts:
\begin{align}
\label{mlp-02}
\begin{split}
    \widetilde{X}_{1} &= \widetilde{X}[:,:,1:c_{in}],\widetilde{X}_{2} =\widetilde{X}[:,:,c_{in}+1:2c_{in}], \\
    \widetilde{X}_{3} &= \widetilde{X}[:,:,2c_{in}+1:3c_{in}]. 
\end{split}
\end{align}
Then, we apply Spatial-Shift for $\mathit{\widetilde{X}}_{1}$ and $\mathit{\widetilde{X}}_{2}$, while $\mathit{\widetilde{X}}_{3}$ keeps the same. Specifically, Spatial-Shift performs the following operations on $\mathit{\widetilde{X}}_{1}$:

\vspace{-10pt}
\begin{align}
\label{mlp-03}
&\scalebox{0.95}{$\widetilde{X}_{1}[2:h,:,1:\frac{c_{in}}{4}]\leftarrow \widetilde{X}_{1}[1:h-1,:,1:\frac{c_{in}}{4}]$},\notag\\
&\scalebox{0.95}{$\widetilde{X}_{1}[1:h-1,:,\frac{c_{in}}{4}+1:\frac{c_{in}}{2}]\leftarrow \widetilde{X}_{1}[2:h,:,\frac{c_{in}}{4}+1:\frac{c_{in}}{2}]$},\notag\\
&\scalebox{0.95}{$\widetilde{X}_{1}[:,2:w,\frac{c_{in}}{2}:\frac{3c_{in}}{4}]\leftarrow \widetilde{X}_{1}[:,1:w-1,\frac{c_{in}}{2}:\frac{3c_{in}}{4}]$},\notag\\
&\scalebox{0.95}{$\widetilde{X}_{1}[:,1:w-1,\frac{3c_{in}}{4}:c_{in}]\leftarrow \widetilde{X}_{1}[:,2:w,\frac{3c_{in}}{4}:c_{in}]$}.\notag\\
\end{align}

For $\widetilde{X}_{2}$, Spatial-Shift conducts:
\begin{align}
\label{mlp-04}
&\scalebox{0.95}{$\widetilde{X}_{2}[:,2:w,1:\frac{c_{in}}{4}]\leftarrow \widetilde{X}_{2}[:,1:w-1,1:\frac{c_{in}}{4}]$},\notag\\
&\scalebox{0.95}{$\widetilde{X}_{2}[:,1:w-1,\frac{c_{in}}{4}+1:\frac{c_{in}}{2}]\leftarrow \widetilde{X}_{2}[:,2:w,\frac{c_{in}}{4}+1:\frac{c_{in}}{2}]$},\notag\\
&\scalebox{0.95}{$\widetilde{X}_{2}[2:h,:,\frac{c_{in}}{2}:\frac{3c_{in}}{4}]\leftarrow \widetilde{X}_{2}[1:h-1,:,\frac{c_{in}}{2}:\frac{3c_{in}}{4}]$},\notag\\
&\scalebox{0.95}{$\widetilde{X}_{2}[1:h-1,:,\frac{3c_{in}}{4}:c_{in}]\leftarrow \widetilde{X}_{2}[2:h,:,\frac{3c_{in}}{4}:c_{in}]$}.\notag\\
\end{align}

To simplify the formulation, we use the $\mathit{S^{2}}(x_{1},x_{2},x_{3})$ to represent Eq.~\eqref{mlp-03} and Eq.~\eqref{mlp-04}. Finally, we combine the outputs from all three branches using element-wise addition and apply split attention, $\mathrm{SA}(x_{1}+x_{2}+x_{3})$ (a mlp operation to recalibrating the importance of different branches) from VIP\cite{tpmai-permutator}. The channel expansion operation from Eq.~\eqref{mlp-01} can be replaced by three standard $ 1\times1 $ convolutions. Thus, we can also decompose this all-MLP paradigm into two steps:

\begin{flalign}
\label{mlp-07}
\begin{split}
    \textbf{Step I}:\widetilde{y}_{i,j}^{(1)}&= K_{(a,b)}^{(1)}x_{i,j},\widetilde{y}_{i,j}^{(2)}=K_{(a,b)}^{(2)}x_{i,j}, \\
    \widetilde{y}_{i,j}^{(3)}&=K_{(a,b)}^{(3)}x_{i,j},
\end{split}&
\end{flalign}

\begin{flalign}
\label{mlp-08}
\textbf{Step II}:&\widetilde{y}_{i,j}^{S(1)},\widetilde{y}_{i,j}^{S(2)},\widetilde{y}_{i,j}^{(3)}= \mathit{S^{2}}(\widetilde{y}_{i,j}^{(1)},\widetilde{y}_{i,j}^{(2)},\widetilde{y}_{i,j}^{(3)}),&&\\
&{y}_{i,j}=\mathrm{MLP}\left(\mathrm{SA}\left(\widetilde{y}_{i,j}^{S(1)}+\widetilde{y}_{i,j}^{S(2)}+\widetilde{y}_{i,j}^{(3)}\right)\right).
\end{flalign}
where $\widetilde{y}_{i,j}^{S}$ denotes the feature map obtained after applying spatial shift to $\widetilde{y}_{i,j}$.
\subsubsection{Unified Search Space}
\label{Unified}
Based on above equations, we successfully design the unified search space for three different types of operators: 
\begin{flalign}
\label{hybrid-1}
\begin{split}
\textbf{Step I}:{q}_{i,j}^{(l)}&=W_{q}^{(l)}x_{i,j}, {k}_{i,j}^{(l)}=W_{k}^{(l)}x_{i,j},\\
{v}_{i,j}^{(l)}&=W_{v}^{(l)}x_{i,j}, 
\end{split}&
\end{flalign}


\begin{flalign}
     \textbf{Step II}&: \notag && \\  
      \text{MSA}& \left\{
    \begin{aligned}
        &\widetilde{y}_{i,j}= \underset{l=1}{\overset{H}{\|}}\mathrm{Attention}\left({q}_{i,j}^{(l)},{k}_{i,j}^{(l)},{v}_{i,j}^{(l)}\right), \\ 
        &{y}_{i,j}= \widetilde{y}_{i,j}W^{O}. 
    \end{aligned} \right. && \\
      \text{Conv}& \left\{
    \begin{aligned}
        &\widetilde{y}_{i,j}^{(a,b)}=\underset{l=1}{\overset{H}{\|}}\left({q}_{i,j}^{(l)},{k}_{i,j}^{(l)},{v}_{i,j}^{(l)}\right)W^{P},\\ 
        &y_{i,j}= \sum_{a,b}{\mathrm{shift}\left(\widetilde{y}_{i,j}^{(a,b)},a,b\right)}. 
    \end{aligned} \right. && \\
      \text{MLP}& \left\{
    \begin{aligned}
    &\widetilde{y}_{i,j}^{S(1)},\widetilde{y}_{i,j}^{S(2)},\widetilde{y}_{i,j}^{(3)}= \mathit{S^{2}}\left({q}_{i,j}^{(l)},{k}_{i,j}^{(l)},{v}_{i,j}^{(l)}\right),\\ 
        &{y}_{i,j}=\mathrm{MLP}\left(\mathrm{SA}\left(\widetilde{y}_{i,j}^{S(1)}+\widetilde{y}_{i,j}^{S(2)}+\widetilde{y}_{i,j}^{(3)}\right)\right). 
    \end{aligned} \right. &&
\end{flalign}

The operations of transformer, convolution, and MLP are completely equivalent in the first step, allowing them to share weights through three $1\times1$ convolutions. In the second step, the network performs different paradigm operations based on the selection.

\subsection{Cross Weight Sharing}
\begin{figure} [t]
    \centering
    \subfloat[Classical weight sharing\label{weight sharing a}]{
        \includegraphics[scale=0.75]{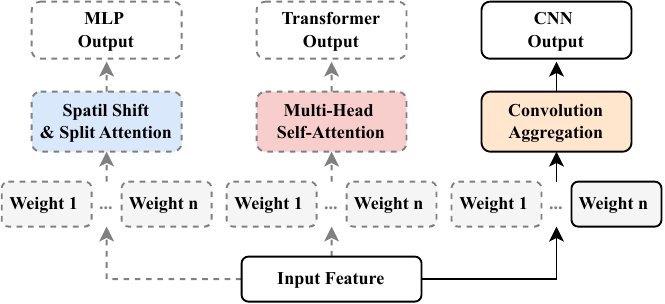}}
        \vspace{-1mm}
    \\
    \subfloat[Cross weight sharing\label{weight sharing b}]{
        \includegraphics[scale=0.75]{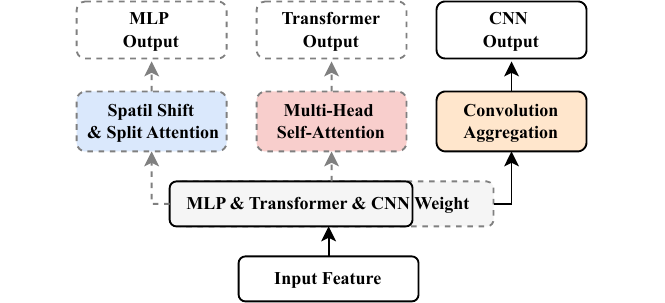} }
    \caption{(a) Classical weight sharing. (b) Cross weight sharing for different types of operators. Each block is depicted as a horizontal bar, with its total length representing the maximum channel width. The solid portion corresponds to the selected active channels, while the dashed portion indicates the unselected ones.}
    \label{hybrid_net} 
    \vspace{-0.6cm}
\end{figure}

One-shot NAS typically share weights among subnets, while maintaining separate weights for different operators in each layer (as illustrated in Fig.~\ref{weight sharing a}). Nevertheless, during training, the number of weight updates is limited. As the number of operators increases, it becomes difficult to update the weight of all blocks, which substantially decreases the efficiency of the architecture search and the performance of subnet, making it impractical for industrial applications. To solve above problems, Autoformer\cite{iccv-autoformer} employs a weight entanglement strategy, allowing different transformer blocks to share weights for their common components in each layer. Nevertheless, such approach is only suitable for pure transformer architectures and does not fit hybrid network architectures. Therefore, we propose a new strategy called cross weight sharing. It consists of two parts. The first part involves weight sharing across different types of operators, achieved by Eq.~\eqref{hybrid-1}. In the second part, we design weight sharing strategies for each type of operator based on their characteristics. In this way, weights can be shared between the same type of operators. With this strategy, all candidate operators' weights can be updated at the same time, as shown in Fig.~\ref{weight sharing b}, thereby accelerating the convergence of the network.

\subsubsection{Convolution}
For convolution, our cross size weight sharing strategy stores weights only for the largest kernel, allowing smaller kernels to directly inherit weights from it. Let $w_{S}^{(i)}$ and $w_{L}^{(i)}$ represent the weights of the small and largest kernels in the $\mathit{i}$-th layer, with $k_{S}$ and $k_{L}$ denoting their respective kernel sizes. The convolution weight shape is $(c_{out}, c_{in}, k, k)$, where $c_{out}$ and $c_{in}$ are the number of output and input channels. Thus, the inheritance relationship of weights can be expressed as:
\begin{equation}
\label{cross_size-1}
    w_{S}^{(i)} = w_{L}^{(i)}[:c_{out},:c_{in}, p:p+k_{S}, q:q+k_{S}]
\end{equation}
where $(p,q)$ denotes the starting position of the process. 
\subsubsection{Transformer and MLP}
For the MLP and transformer, we similarly store the weights of the largest blocks. Since the basic components of Transformers and MLPs are fully connected layers, their size mainly depends on the embedding dimension (input channels) and output channels. The shape of the weight can be expressed as $(c_{out}, c_{in})$, Therefore, the cross weight sharing strategy for them can be simply defined as:
\begin{equation}
\label{cross_size-2}
    w_{S}^{(i)} = w_{L}^{(i)}[:c_{out},:c_{in}]
\end{equation}
where $w_{S}^{(i)}$ and $w_{L}^{(i)}$ are the weights of the small and largest blocks, respectively.

\subsection{Multi-level Feature Aggregation Module}
\label{mlfg search space}
Aggregating multi-level features has been proven effective for surface defect detection\cite{tim-global_local,tim-Deep-Feature}. Instead of using manually designed architectures, we develop a searchable feature aggregation module, illustrated in the bottom right corner of Fig.~\ref{network}. This module, inspired by the FPN\cite{cvpr-fpn} and UperNet\cite{eccv-upernet}, refines multi-level features using a series of feature extraction operations. The whole process is designed as a directed acyclic graph with $\mathit{N}_{F}$ nodes and $\mathit{N}_{E}$ edges. Each node represents a feature map, and each directed edge corresponds to one of five candidate operators (1×1, 3×3, 5×5, dilated 3×3, depthwise 3×3 convolution).

After feature refinement, a Pyramid Pooling Module (PPM) from PSPNet\cite{cvpr-psp} is applied on the last layer of the backbone network. The pooling size is also searchable, ranging from 1 to 6. Then, the multi-level features are fused together to produce the final output. The joint search over the backbone and MFAM allows adaptive selection of fusion strategies alongside feature extraction, enabling flexible refinement of both low and high level features.

\begin{table}[t]
\centering
\caption{Search space of the encoder. Tuples of three values in parentheses represent the lowest value, highest, and steps.}
\label{search space}
\begin{tabular}{c|c|c|c|c}
\specialrule{0.8pt}{0pt}{0pt}
Stage & Depth                      & Channel        & FFN ratio                        & Operator                                                                                   \\ \hline
1     & \multirow{4}{*}{(1, 4, 1)} & (32, 64, 32)   & \multirow{2}{*}{(7.5, 9.0, 0.5)} & \multirow{4}{*}{\begin{tabular}[c]{@{}c@{}}Transformer,\\ Convolution,\\ MLP\end{tabular}} \\ 
2     &                            & (32, 128, 32)  &                                  &                                                                                            \\ 
3     &                            & (128, 224, 32) & \multirow{2}{*}{(3.5, 5.0, 0.5)} &                                                                                            \\ 
4     &                            & (192, 288, 32) &                                  &                                                                                            \\ 
\specialrule{0.8pt}{0pt}{0pt}
\end{tabular}
\end{table}

\begin{figure}[t]
    \centering
    \includegraphics[width=0.489\textwidth]{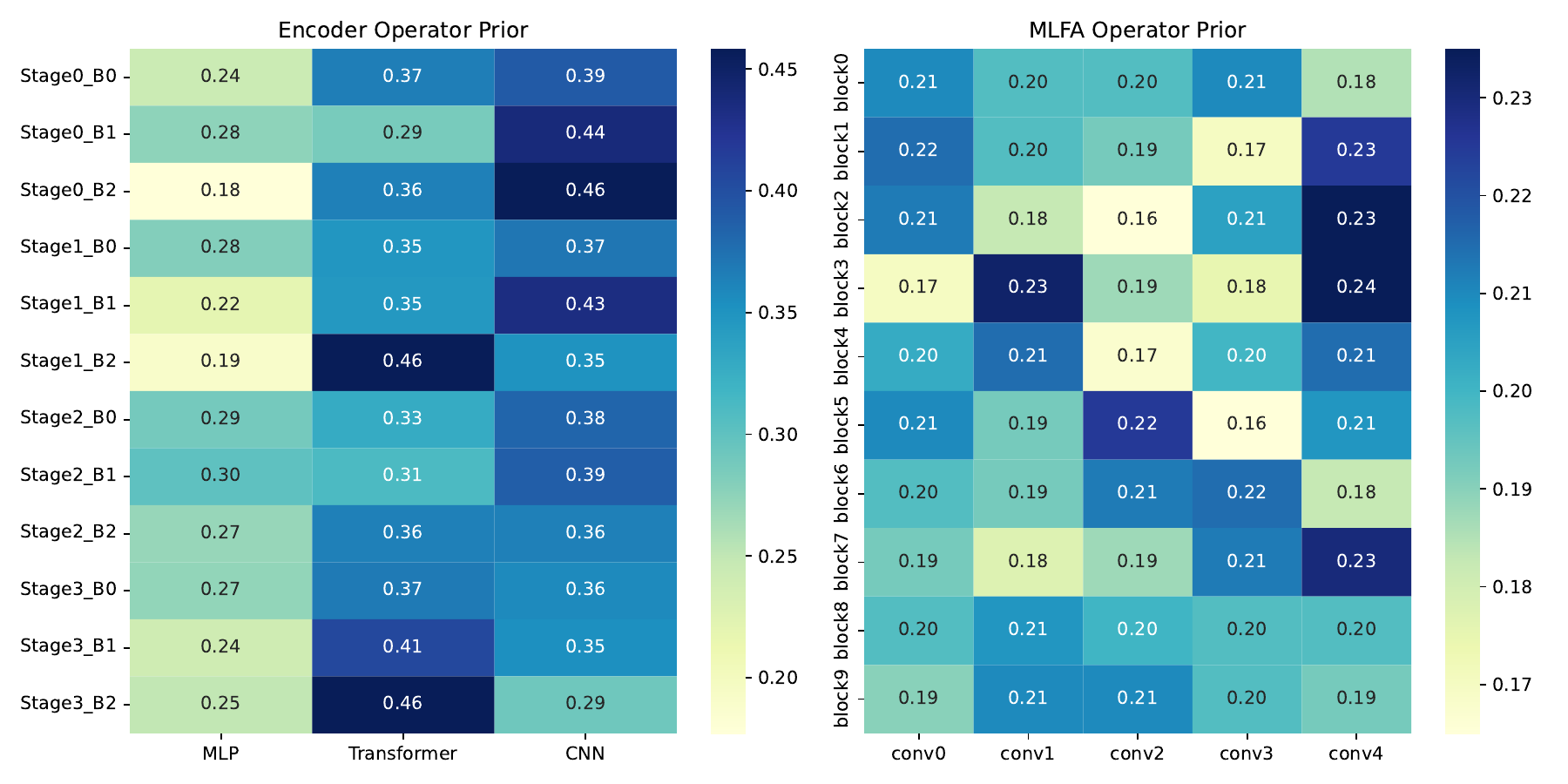} 
    \caption{Latency-aware prior constructed from runtime statistics during supernet training. Each cell corresponds to a specific operator at a specific block location, with warmer colors indicating lower average latency and higher sampling probability.}
    \label{prior}
    \vspace{-0.6cm}
\end{figure}





\begin{algorithm}
\caption{Evolution Search with Latency-Aware Prior}
\label{alg:evolutionary_search}
\begin{algorithmic}[h]
    \REQUIRE Trained supernet $\mathcal{S}$, search space $\mathcal{A}$, latency prior $\mathcal{P}_{\text{lat}}$, generations $G$, population size $N$, mutation rate $p_m$, crossover rate $p_c$, validation set $\mathcal{D}_{val}$, score weight $\lambda$
    \ENSURE Optimal subnet $\alpha^*$

    \STATE Initialize operator prior $\mathcal{P}_{\text{lat}}$ from low-latency subnets
    \STATE Initialize population $\mathcal{P}$ via $\mathcal{P}_{\text{lat}}$-biased sampling from $\mathcal{A}$
    \STATE Define score function: $\mathrm{Score}(\alpha)=\mathrm{mIoU}(\alpha)-\lambda \cdot \mathrm{Latency}(\alpha)$

    \FOR{$t = 1$ to $G$}
        \STATE Evaluate score for each $\alpha \in \mathcal{P}$ on $\mathcal{D}_{val}$
        \STATE $\mathcal{P}_{\text{parent}} \leftarrow$ Select top-$k$ subnets from $\mathcal{P}$ by score
        \STATE $\mathcal{P}_{\text{child}} \leftarrow \emptyset$
        \IF{Random() $< p_m$}
            \STATE $\mathcal{P}_{\text{child}} \leftarrow$ Mutation($\mathcal{P}_{\text{parent}}$) biased by $\mathcal{P}_{\text{lat}}$
        \ENDIF
        \IF{Random() $< p_c$}
            \STATE $\mathcal{P}_{\text{child}} \leftarrow$ Crossover($\mathcal{P}_{\text{parent}}$)
        \ENDIF

        \STATE $\mathcal{P} \leftarrow \mathcal{P} \cup \mathcal{P}_{\text{child}}$
    \ENDFOR

    \STATE $\alpha^* \leftarrow \arg\max_{\alpha \in \mathcal{P}} \mathrm{Score}(\alpha)$
    \RETURN $\alpha^*$
\end{algorithmic}
\end{algorithm}

\subsection{Evolution Search with Latency-Aware Prior}
Once the supernet has fully converged, we employ an evolutionary algorithm to search for the optimal subnet. To guide the search toward deployable architectures, we construct a latency-aware prior based on runtime statistics collected during supernet training. Specifically, during each training iteration, a candidate subnet is sampled and executed on the target hardware, and its inference latency is recorded. Over the course of training, we accumulate these latency records for each type of operator in each searchable block. Based on this data, we estimate the average latency of each operator-block pair, forming a block-level latency map.

This latency map is then converted into a probabilistic prior, where operators with lower latency are assigned higher sampling probabilities. We visualize this prior as a heatmap in Fig.~\ref{prior}, where warmer colors indicate faster operators favored during search. The latency-aware prior is used in two key steps: (1) it guides the initialization of the population by increasing the likelihood of selecting fast operators, and (2) it biases the mutation process such that operators with lower average latency have a higher probability of being selected during mutation. The complete procedure of latency-aware evolution Search is outlined in Algorithm~\ref{alg:evolutionary_search}.

\section{Experiments}

\subsection{Datasets}
We evaluate our AutoNAD on three different industrial defect datasets, which encompass most types of surface defects.

\textbf{NEU dataset}\cite{SONG2013858} is a hot-rolled steel strip surface defect dataset containing three typical defects: inclusion, patches, and scratches. Each type of defect includes 300 images, all with a resolution of $ 200\times200 $ pixels.

\textbf{MSD dataset}\cite{FDSNeT} comprises 1,200 images of three types of defects: oil, stain, and scratch on the surface of mobile phone screens. The original resolution is $1920\times1080$ pixels, and we resized them to $960\times540$ pixels during training and testing.

\textbf{MT dataset}\cite{Magnetic} contains 1,344 images with five types of defects: blowhole, break, crack, fray, and uneven. Most samples have different resolutions. For training and testing, we selected only the defect images (782 images) from MT dataset and resized all the images to $ 320\times320 $ pixels.

\begin{table*}[ht]
\centering
‌\footnotesize‌
\tabcolsep=0.11cm
\caption{Performance comparisons on the NEU, MT, and MSD datasets.}
\label{Per Com}
\begin{tabular}{l|cc|c|ccc|ccc|ccc}
\specialrule{0.8pt}{0pt}{0pt}
\multirow{2}{*}{Method} & \multirow{2}{*}{\begin{tabular}[c]{@{}c@{}}Encoder\\ Type\end{tabular}} & \multirow{2}{*}{\begin{tabular}[c]{@{}c@{}}Design\\ Type\end{tabular}} & \multirow{2}{*}{\begin{tabular}[c]{@{}c@{}}Latency\\ (ms)\end{tabular}} & \multicolumn{3}{c|}{NEU}                           & \multicolumn{3}{c|}{MT}                            & \multicolumn{3}{c}{MSD}                            \\ \cline{5-13} 
                        &                                                                         &                                                                        &                                                                         & \multicolumn{1}{c|}{\#Params} & mIoU(\%) & mF1(\%) & \multicolumn{1}{c|}{\#Params} & mIoU(\%) & mF1(\%) & \multicolumn{1}{c|}{\#Params} & mIoU(\%) & mF1(\%) \\ \hline
PSPNet $_\text{2017}$~\cite{cvpr-psp}              & C                                                                       & Manual                                                                 & 10.8                                                                    & \multicolumn{1}{c|}{51.1M}    & 83.8     & 91.0    & \multicolumn{1}{c|}{51.1M}    & 78.8     & 87.2    & \multicolumn{1}{c|}{51.1M}    & 89.6     & 94.3    \\
Deeplabv3+ $_\text{2018}$~\cite{eccv-deeplabv3_plus}         & C                                                                       & Manual                                                                 & 10.4                                                                    & \multicolumn{1}{c|}{59.2M}    & 84.1     & 91.1    & \multicolumn{1}{c|}{59.2M}    & 74.9     & 85.2    & \multicolumn{1}{c|}{59.2M}    & 88.8     & 93.8    \\
AutoDeepLab $_\text{2019}$~\cite{cvpr-Auto-DeepLab}         & C                                                                       & Auto                                                                   & 35.5                                                                    & \multicolumn{1}{c|}{14.2M}    & 76.8     & 86.1    & \multicolumn{1}{c|}{12.8M}    & 71.0     & 80.6    & \multicolumn{1}{c|}{18.5M}    & 91.0     & 95.1    \\
FasterSeg $_\text{2020}$~\cite{fasterseg}          & C                                                                       & Auto                                                                   & 5.1                                                                     & \multicolumn{1}{c|}{3.6M}     & 76.5     & 86.0    & \multicolumn{1}{c|}{4.2M}     & 73.0     & 82.8    & \multicolumn{1}{c|}{4.2M}     & 83.6     & 90.4    \\
HR-NAS $_\text{2021}$~\cite{CVPR-hr_nas}             & H                                                                       & Auto                                                                   & 36.3                                                                    & \multicolumn{1}{c|}{14.1M}    & 83.4     & 90.6    & \multicolumn{1}{c|}{14.1M}    & 78.5     & 87.5    & \multicolumn{1}{c|}{14.1M}    & 90.7     & 95.0    \\
Swim-T $_\text{2021}$~\cite{ICCV-swin}            & T                                                                       & Manual                                                                 & 18.7                                                                    & \multicolumn{1}{c|}{57.1M}    & 83.9     & 91.0    & \multicolumn{1}{c|}{57.1M}    & 78.0     & 87.0    & \multicolumn{1}{c|}{57.1M}    & 90.1     & 94.7    \\
Segformer $_\text{2021}$~\cite{nips-segformer}          & T                                                                       & Manual                                                                 & 31.3                                                                    & \multicolumn{1}{c|}{64.0M}    & 83.6     & 90.8    & \multicolumn{1}{c|}{64.0M}    & 71.6     & 81.9    & \multicolumn{1}{c|}{64.0M}    & 88.9     & 93.9    \\
Topformer $_\text{2022}$~\cite{cvpr-topformer}          & T                                                                       & Manual                                                                 & 6.5                                                                     & \multicolumn{1}{c|}{5.1M}     & 78.9     & 87.7    & \multicolumn{1}{c|}{5.1M}     & 71.8     & 82.0    & \multicolumn{1}{c|}{5.1M}     & 81.3     & 89.1    \\
DDRNet39 $_\text{2023}$~\cite{ddrnet}         & C                                                                       & Manual                                                                 & 8.2                                                                     & \multicolumn{1}{c|}{32.3M}    & 82.8     & 90.3    & \multicolumn{1}{c|}{32.3M}    & 79.4     & 87.9    & \multicolumn{1}{c|}{32.3M}    & 82.6     & 90.2    \\
PIDNet-M $_\text{2023}$~\cite{CVPR_pidnet}           & C                                                                       & Manual                                                                 & 7.2                                                                     & \multicolumn{1}{c|}{34.4M}    &82.1          &89.8         & \multicolumn{1}{c|}{34.4M}    &78.8          &87.5         & \multicolumn{1}{c|}{34.4M}    &82.3          &90.0        \\
CycleMLP $_\text{2023}$~\cite{tpami-cycle-mlp}           & M                                                                       & Manual                                                                 & 38.9                                                                    & \multicolumn{1}{c|}{55.6M}    & 81.9     & 89.7    & \multicolumn{1}{c|}{55.6M}    & 77.6     & 86.8    & \multicolumn{1}{c|}{55.6M}    & 85.4     & 91.7    \\

SDPT $_\text{2024}$~\cite{SDPT-tits}       & T                                                                       & Manual                                                                 & 15.4                                                                        & \multicolumn{1}{c|}{28.6M}         &82.1          &89.9         & \multicolumn{1}{c|}{28.6M}         &78.4          &87.2         & \multicolumn{1}{c|}{28.6M}         &86.9          &92.6         \\
EfficientSAM $_\text{2024}$~\cite{CVPR-efficient-sam}      & T                                                                       & Manual                                                                 &66.7                                                                         & \multicolumn{1}{c|}{10.2M}         &72.0          &83.1         & \multicolumn{1}{c|}{10.2M}         &72.3         &83.6         & \multicolumn{1}{c|}{10.2M}         &45.0          &47.2         \\
DSNet $_\text{2025}$~\cite{tcsvt-dsnet}                & C                                                                       & Manual                                                                 &9.8                                                                        & \multicolumn{1}{c|}{6.8M}         &83.0          &90.4         & \multicolumn{1}{c|}{6.8M}         &80.0          &88.5         & \multicolumn{1}{c|}{6.8M}         &85.6          &91.9         \\ \hline
\textbf{AutoNAD (Ours) }             & H                                                                       & Auto                                                                   &8.3                                                                         & \multicolumn{1}{c|}{6.3M}         &\textbf{84.6}          &\textbf{91.4}         & \multicolumn{1}{c|}{4.6M}         &\textbf{81.3}         &\textbf{89.3}         & \multicolumn{1}{c|}{5.9M}         &\textbf{92.5}          &\textbf{96.0}         \\
\specialrule{0.8pt}{0pt}{0pt}
\end{tabular}
\end{table*}

\subsection{Experimental Setup}
\subsubsection{Evaluation Metrics}
To evaluate the performance of our framework, mean intersection over union (mIoU) and mean F1 score (mF1) are used as the primary metrics. IoU represents the ratio of the intersection area to the union area between the predicted results and the ground truth. F1 is the harmonic mean of precision and recall.
\subsubsection{Implementation Details}
The pipeline of our method is divided into three stages:

\textbf{Supernet training:} For all datasets, we use AdamW with an initial learning rate of 0.0005 and cosine decay. Standard augmentations (random flip, resize, rotation, etc.) are applied. The number of epochs is set to 800 (NEU), 2000 (MSD), and 3000 (MT).

\textbf{Evolution Search:} We set the population $\mathcal{P}$ to 100 and the number of generations $\mathit{G}$ to 50. The top 10 subnets are selected as the parental models in each generation. The mutation probability $p_m$ and crossover $p_c$ are set to 0.2 and 0.4.

\textbf{Subnet retraining:} For all datasets, we use Adam with an initial learning rate of 0.0001 and the poly decay scheduler. The augmentation strategy remains consistent with the supernet, and only cross-entropy loss is used. The retraining epoch is set to 500 for all datasets.

Note that all the experiments are implemented on NVIDIA GeForce RTX 3090 GPUs.
\subsection{Performance Comparison}
We select several mainstream SOTA methods to compare against our approach, including real-time detection method~\cite{CVPR_pidnet,ddrnet,tcsvt-dsnet,cvpr-topformer,fasterseg} and other representative classical models~\cite{cvpr-psp,eccv-deeplabv3_plus,cvpr-Auto-DeepLab,CVPR-hr_nas,ICCV-swin,tpami-cycle-mlp,SDPT-tits,nips-segformer}. In addition, as foundation models have achieved remarkable success in vision tasks, we also compare our method with large-scale models~\cite{CVPR-efficient-sam}. We follow the setting in \cite{CVPR-efficient-sam}, using box and point as prompts. The performance comparisons on NEU, MT, and MSD datasets are shown in Table~\ref{Per Com}. Note that the latency metrics are measured on the MT dataset. Here, C, T, M, and H denote CNN, Transformer, MLP, and Hybrid architectures, respectively.

On NEU dataset, AutoNAD achieves the best performance, with the mIoU and F1 score reaching 84.6\% and 91.4\%. It outperforms DeepLabv3+~\cite{eccv-deeplabv3_plus} by 0.5\% in mIoU and 0.3\% in F1, despite having nearly 10 times fewer parameters. On MT dataset, AutoNAD attains the best performance of 81.3\% in mIoU and 89.3\% in F1, with an improvement of 1.3\% in mIoU and 0.8\% in F1 compared to DSNet~\cite{tcsvt-dsnet}. On MSD dataset, the proposed model achieves an mIoU improvement of 1.5\% over AutoDeeplab (auto-designed)~\cite{cvpr-Auto-DeepLab}. In addition, our model maintains a relatively low inference latency, making it more suitable for deployment on detection equipment in industrial production lines. 

We also observed that large foundation models perform relatively poorly on industrial defect datasets. This is mainly because they are pre-trained on natural images and lack domain-specific knowledge for industrial applications. Moreover, detection based on large models still relies on manually crafted prompts, which contradicts the requirement for automation in real-world industrial production. 
\subsection{Search Efficiency}
In this part, we evaluate the search efficiency of different NAS methods on NEU dataset. As shown in Table~\ref{search cost}, our method requires substantially less search time compared to other NAS methods (the reported search time includes both supernet training and the subsequent evolutionary search). This demonstrates that AutoNAD not only achieves the best overall performance, but also offers superior search efficiency.

\subsection{Model generalization}
Model generalization is critical in industrial quality inspection. We apply the subnets searched on specific datasets to train and test on other unseen datasets. As shown in Table~\ref{cross_validation}, our AutoNAD achieves consistently strong results across datasets. It outperforms existing methods in overall performance and demonstrates excellent transferability.

\begin{table}[t]
\caption{Search Cost Comparison with different NAS methods}
\label{search cost}
\centering
\tabcolsep=0.5cm
‌\scriptsize
\begin{tabular}{c|c}
\specialrule{0.8pt}{0pt}{0pt}
Method & \makecell{Search Cost (GPU Hours)}  \\ \hline
HR-NAS\cite{CVPR-hr_nas}         &9.24                                                 \\
FasterSeg\cite{fasterseg}       & 12.00                                                 \\
AutoDeepLab\cite{cvpr-Auto-DeepLab}     & 15.60                                      \\ \cline{1-2}
\textbf{AutoNAD}       & \textbf{1.60}                                                  \\ 
\specialrule{0.8pt}{0pt}{0pt}
\end{tabular}
\end{table}

\begin{table}[t]
\caption{Comparison of models searched on different datasets}
\label{cross_validation}
\centering
\scriptsize
\tabcolsep=0.1cm
\begin{tabular}{c|cc|cc|cc}
\specialrule{0.8pt}{0pt}{0pt}
\multirow{2}{*}{Method} & \multicolumn{2}{c|}{NEU} & \multicolumn{2}{c|}{MT} & \multicolumn{2}{c}{MSD} \\ \cline{2-7} 
                        & mIoU(\%)    & mF1(\%)    & mIoU(\%)   & mF1(\%)    & mIoU(\%)   & mF1(\%)    \\ \hline
\makecell{AutoNAD (NEU)}  &\textbf{84.6}    &\textbf{91.4}   &80.7   &88.6   &92.3   &95.9   \\
\makecell{AutoNAD (MT)}   &84.4    &91.3   &\textbf{81.3}   &\textbf{89.3}   &92.0   &95.8   \\
\makecell{AutoNAD (MSD)}  &84.3    &91.3   &80.8   &88.7   &\textbf{92.5}   &\textbf{96.0}   \\ 
\specialrule{0.8pt}{0pt}{0pt}
\end{tabular}
\end{table}

\begin{table}[!t]
\centering
\scriptsize
\caption{Comparison of Different Search Spaces on MT dataset}
\label{Unified Seach Space}
\tabcolsep=0.4cm
\begin{tabular}{ccc|c|c}
\specialrule{0.8pt}{0pt}{0pt}
\multicolumn{1}{c}{CNN}    & Transformer   & MLP    &mIoU(\%)   & mF1(\%) \\ \hline
\checkmark                           &    &                        &79.7                                                      &88.2                                     \\
                          &\checkmark      &                        &79.9                                                        &88.4                                     \\
                          &     &\checkmark                        &79.5                                                       &88.1                                     \\
\checkmark                          &\checkmark     &                        &80.3                                                       &88.6                                     \\
\checkmark                        &     &\checkmark                        &80.7                                                       & 88.9                                   \\
                        &\checkmark     &\checkmark                        & 80.9                                                      & 89.1                                    \\
 \checkmark                         & \checkmark    &\checkmark     &\textbf{81.3}                                     &\textbf{89.3}                                                                         \\ 
\specialrule{0.8pt}{0pt}{0pt}
\end{tabular}
\end{table}

\subsection{Ablation Study}
\subsubsection{Impact of Unified Seach Space}
We first conduct ablation studies to verify the effectiveness of our proposed unified search space, with the results presented in Table~\ref{Unified Seach Space}. The experiment investigates a total of seven distinct search spaces. Among them, combination involving all three types of operators (CNN, transformer, and MLP) achieves the best performance, with an mIoU of 81.3\% and an mF1 of 89.3\%. In comparison, combinations such as CNN with Transformer or CNN with MLP yield slightly lower results. Additionally, models derived from single-operator search spaces exhibit the poorest performance. The experimental results indicate that combining different types of operators leads to more accurate detection when addressing diverse surface defects, underscoring the effectiveness of our unified search space.

\begin{table}[t]
\caption{Comparison of Different Decoders}
\label{MFAM}
\tabcolsep=0.10cm
\scriptsize
\centering
\begin{tabular}{c|cc|cc|cc}
\specialrule{0.8pt}{0pt}{0pt}
\multirow{2}{*}{Decoder} & \multicolumn{2}{c|}{NEU}                     & \multicolumn{2}{c|}{MT}                      & \multicolumn{2}{c}{MSD}                     \\ \cline{2-7} 
                         &mIoU(\%)              &mF1(\%)                &mIoU(\%)             &mF1(\%)                 &mIoU(\%              &mF1(\%))               \\ \hline
UperNet\cite{eccv-upernet}                  &83.6         &90.8          &77.0             &86.3              &89.8             &94.5             \\
Semantic FPN\cite{cvpr-fpn}             &73.2         &83.7          & 73.4            &83.8              &88.4             &93.7            \\
Segformer\cite{nips-segformer}                &84.2         &91.2          & 70.0            &81.1              &86.7             &92.5             \\
AutoNAD (Ours)                  &\textbf{84.6}             & \textbf{91.4}             &\textbf{81.3}             & \textbf{89.3}             &\textbf{92.5}            &\textbf{96.0}             \\
\specialrule{0.8pt}{0pt}{0pt}
\end{tabular}
\end{table}

\begin{table}[!t]
\caption{Comparison of Different Supernet Training Methods on NEU Dataset}
\label{Cross Weight Sharing}
\tabcolsep=0.55cm
‌\scriptsize
\centering
\begin{tabular}{c|cc}
\specialrule{0.8pt}{0pt}{0pt}
\multirow{2}{*}{Method} & \multicolumn{2}{c}{mIoU (\%)} \\
\cline{2-3}
 & Inherited & Retrain \\
\hline
Classical Weight Sharing   & 62.1 & 83.9 \\
Weight Entangle~\cite{iccv-autoformer}  & 72.5 & 84.2 \\
Cross Weight Sharing  & \textbf{79.0} & \textbf{84.6} \\
\specialrule{0.8pt}{0pt}{0pt}
\end{tabular}
\end{table}

\subsubsection{Impact of MFAM}
In this experiment, we explore the impact of our MFAM. For a fair comparison, we fixed the backbone of the model as our AutoNAD and replaced only the encoder with several manually designed methods (UperNet~\cite{eccv-upernet}, Semantic FPN~\cite{cvpr-fpn} and Segformer~\cite{nips-segformer}), all of which also adopt a multi-level feature fusion strategy. As shown in Table~\ref{MFAM}, Segformer achieves competitive results on NEU dataset, with the mIoU and F1 score reaching 84.2\% and 91.2\%, respectively. However, it fails to accurately detect the defects on MT dataset, showing an mIoU 11.3\% lower than our method. UperNet and Semantic FPN both have relatively stable performance across the three datasets, but the overall accuracy is considerably lower than that of our method. These results owing to their lack of adjustments for the characteristics of industrial defects in feature fusion and feature refinement operations. In contrast, our method is trained on industrial datasets and can search for the most suitable multi-level feature fusion approaches for industrial defects, thus achieving optimal accuracy across various datasets.
\begin{figure}[!t]
    \centering
    \includegraphics[width=0.488\textwidth]{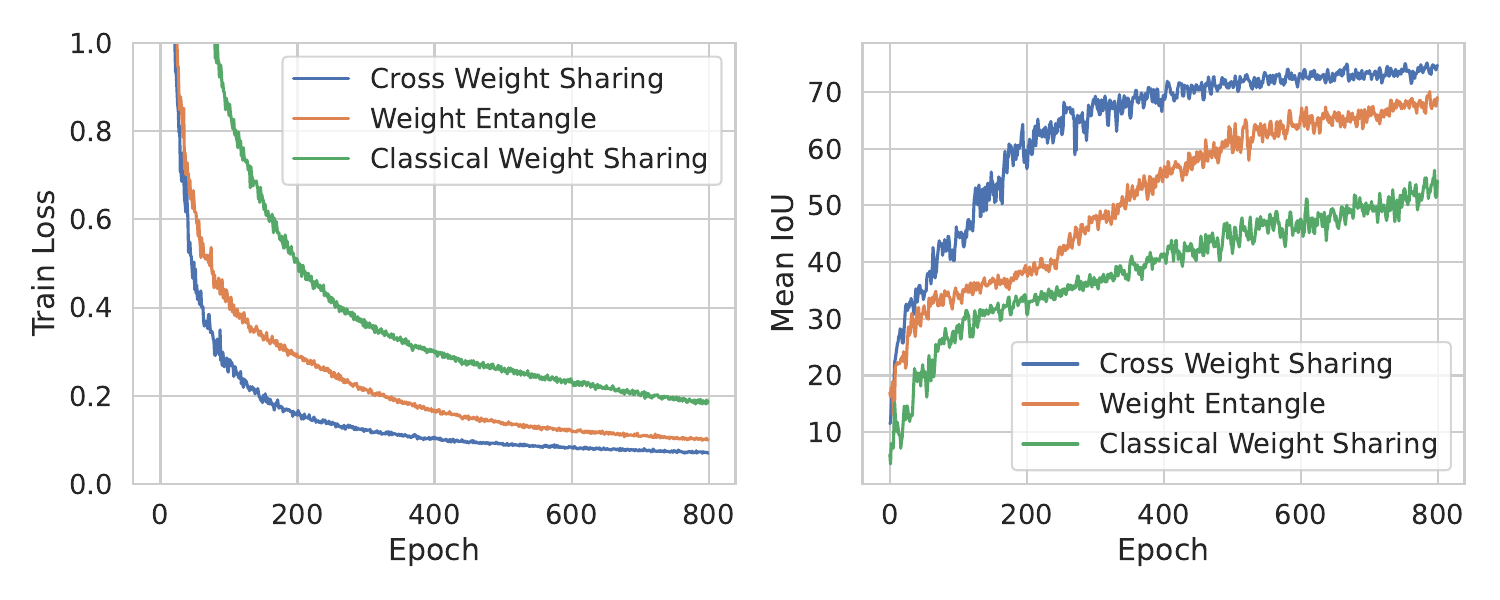} 
    \caption{Left: Comparison of training loss for the supernet on NEU dataset. Right: Comparison of mIoU for subnets during supernet training on NEU dataset.}
    \label{loss_acc}
    \vspace{-0.4cm}
\end{figure}
\subsubsection{Impact of Cross Weight Sharing}
We evaluate cross weight sharing against two baselines: (1) classical weight sharing (each operator independent), and (2) weight entanglement~\cite{iccv-autoformer}, which in our implementation is generalized beyond Transformer blocks so that all operators of the same type share weights within each layer. All methods adopt the same search space and evolution strategy.

As shown in Fig.~\ref{loss_acc}, cross weight sharing achieves the fastest convergence. Classical weight sharing updates each operator independently, leading to low update frequency and slower convergence. Weight entanglement improves efficiency for homogeneous operators but cannot share weights across different types of operators. Our method constructs a unified search space with tailored sharing strategies for different operator types, allowing shared parameters in common parts. This ensures more comprehensive weight updates during training, improving both convergence and performance. Besides, in Table~\ref{Cross Weight Sharing}, subnets inherited directly from the supernet achieve higher mIoU using cross weight sharing, indicating that our method can fully exploit the potential of sampled subnets and facilitate the discovery of the optimal subnet.

Regarding the effectiveness of cross weight sharing, the more detailed reasons are summarized as follows: (1) \textbf{Fair optimization:} Equalizing update opportunities for all subnets prevents bias toward faster-converging candidates~\cite{ICCV_FairNAS}. (2) \textbf{Implicit regularization:} Sampling small subnets discards portions of hidden units~\cite{iccv-autoformer}, similar to dropout~\cite{dropout}, encouraging generalization. For convolutions, smaller kernels inherit from larger ones, retaining receptive fields while avoiding unnecessary complexity.



\begin{figure*}[t]
    \centering
    \includegraphics[width=1.0\textwidth]{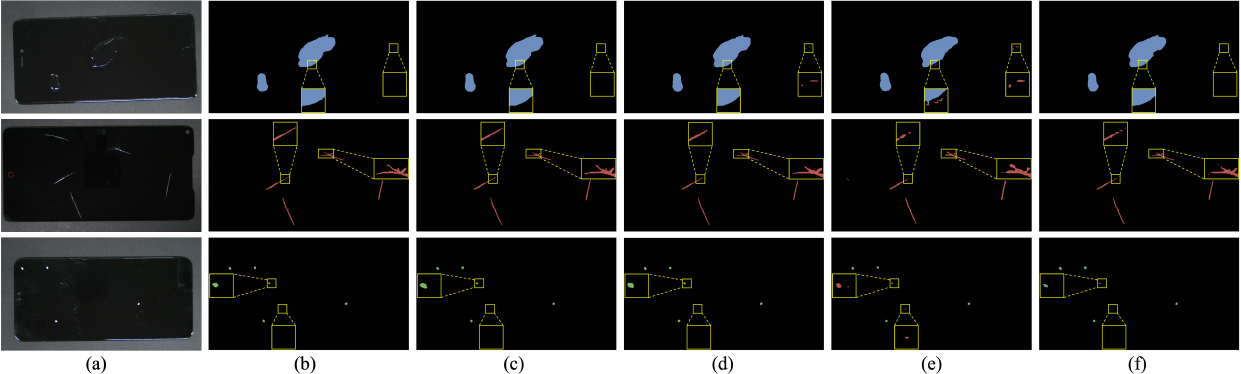} 
    \caption{The visualizations of detection results for different methods on MSD dataset. (a) Original image. (b) GT. (c) AutoNAD. (d) DeepLabv3+. (e) CycleMLP. (f) Segformer.}
    \label{msd}
\end{figure*}

\begin{figure}[t]
    \centering
    \includegraphics[width=0.489\textwidth]{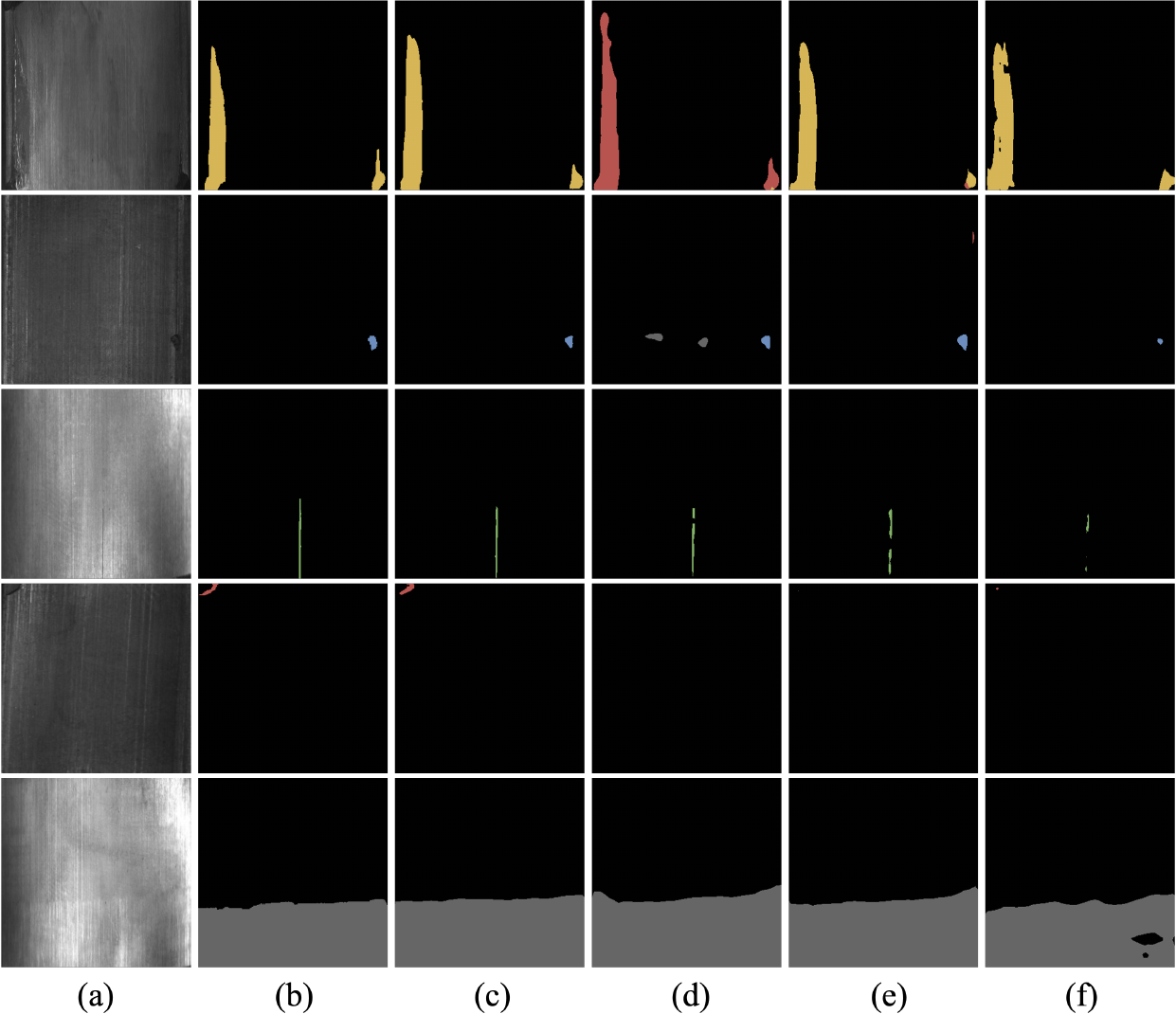} 
    \caption{The visualizations of detection results for different methods on MT dataset. (a) Original image. (b) GT. (c) AutoNAD. (d) DeepLabv3+. (e) CycleMLP. (f) Segformer.}
    \label{mt}
    \vspace{-0.55cm}
\end{figure}

\begin{figure}[t]
    \centering
    \includegraphics[width=0.489\textwidth]{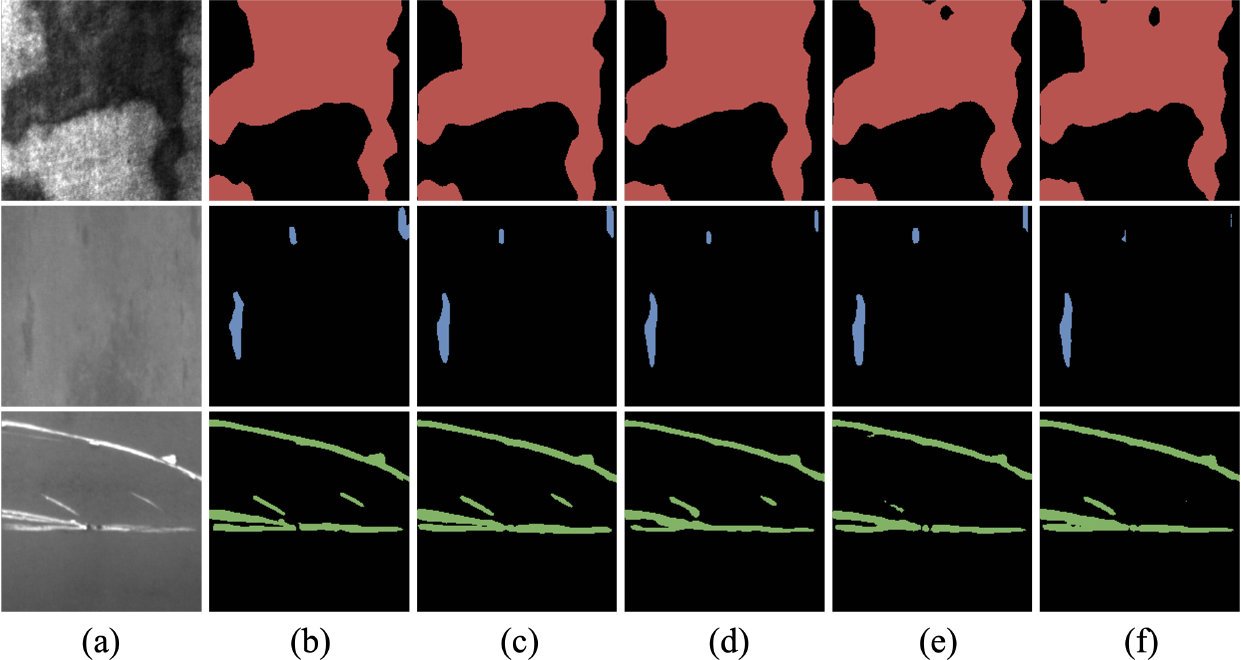} 
    \caption{The visualizations of detection results for different methods on NEU dataset. (a) Original image. (b) GT. (c) AutoNAD. (d) DeepLabv3+. (e) CycleMLP. (f) Segformer.}
    \label{neu}
    \vspace{-0.3cm}
\end{figure}
\subsection{Visualization}
In this section, we present defect detection results of several representative methods on three datasets, including CNN-based methods, transformer-based methods, and MLP-based methods. As shown in Fig.~\ref{neu} (Rows 2 and 3), Segformer~\cite{nips-segformer} exhibits insufficient sensitivity to local details, leading to missed detections when facing the issue of intraclass difference. Although Deeplabv3+~\cite{eccv-deeplabv3_plus} (a CNN-based method) can alleviate this problem, it fails to distinguish between different categories with similar features, as evidenced in Fig.~\ref{mt} (Row 1), where it incorrectly identifies the fray defect as a break. While the MLP-based method performs relatively well on the training set, it may exhibit severe recognition errors when generalized to the test set, as illustrated in Fig.~\ref{msd} (Rows 1 and 3). In contrast, our AutoNAD framework achieves excellent inference results in all the aforementioned situations, demonstrating significant superiority.

\subsection{Industrial Application Validation}
\label{application}
To validate the practical effectiveness of our AutoNAD, we collaborated with an aero engine corporation to develop an automated platform (Fig.~\ref{platform}) for detecting surface defects on aero-engine blades. This platform consists of two parts: imaging and detection. The imaging system includes a 6-DOF robotic arm and a 5-DOF motion platform, along with high-precision cameras, high-resolution lenses, and multispectral light sources. The detection part comprises a GPU computing platform and software that integrates our auto-designed detection network. 

The workflow of this platform can be described in three steps. First, the robotic arm loads the blade onto the 5-DOF motion platform. This motion mechanism, combined with the camera, performs a full-coverage scan and imaging of the blade surface using an integrated path-planning algorithm. Then, the generated defect dataset is sent to the detection system for analysis. During the first analysis phase, our AutoNAD automatically designs a high-accuracy, deployment-friendly detection model based on the collected data and the available computing platform. This model is then used for subsequent defect detection. After detection, the robotic arm unloads the blade for further processing. 

We conducted extensive testing using images collected from an actual production line. As shown in Table~\ref{aero_results}, AutoNAD achieves 84.8\% mIoU with fast inference (8.5 ms on RTX 3090; 85.1 ms on Jetson Xavier NX), demonstrating its effectiveness and adaptability for deployment on edge devices. These results confirm its potential for integration into intelligent manufacturing systems to improve quality control.

Moreover, in practical industrial deployment, the architecture search of AutoNAD is typically a one-time cost. Once an optimal architecture has been discovered, it can be reused across multiple production batches and related product lines, with only lightweight retraining required to adapt to minor data shifts. A new search is usually triggered only when the visual appearance of defects changes significantly or when hardware constraints are updated.

\begin{table}[t]
\centering
\caption{Performance comparison on real production line.}
\scriptsize
\label{aero_results}
\tabcolsep=0.23cm
\begin{tabular}{c|c|cc|c}
\specialrule{0.8pt}{0pt}{0pt}
Method & Params & Latency@3090 & Latency@Jetson & mIoU\% \\ 
\hline
Segformer~\cite{nips-segformer}       & 64.0M  & 31.5  & 378.0   & 55.8 \\
CycleMLP~\cite{tpami-cycle-mlp}       & 55.6M  & 39.1  & 351.9   & 67.6 \\
AutoDeepLab~\cite{CVPR-hr_nas}        & 12.8M  & 36.6  & 310.2   & 72.0 \\
HR-NAS~\cite{CVPR-hr_nas}             & 14.1M  & 38.0  & 323.9   & 83.1    \\
Deeplabv3+~\cite{eccv-deeplabv3_plus} & 59.2M  & 10.9  & 110.8   & 77.4 \\
DDRNet39~\cite{ddrnet}                & 32.3M  & 8.3   & 80.9    & 78.4 \\
SDPT~\cite{SDPT-tits}                 & 28.6M  & 15.7  & 130.1   & 82.5    \\
DSNet~\cite{tcsvt-dsnet}              & 6.8M   & 9.8   & 101.6   & 73.3 \\ \hline
\textbf{AutoNAD}                      & 5.9M   & 8.5   & 85.1    & \textbf{84.8} \\ 
\specialrule{0.8pt}{0pt}{0pt}
\end{tabular}
\end{table}


\IEEEpubidadjcol
\section{Conclusion}
In this paper, we propose AutoNAD, an automated architecture design framework for surface defect detection. The method focuses on addressing two fundamental challenges: intraclass difference and interclass similarity. To this end, AutoNAD performs a unified search over convolution, transformer, and MLP operators, enabling balanced modeling of local details and global context. A cross-operator weight sharing strategy is introduced to enhance training efficiency and improve the quality of searched subnets. In addition, a searchable multi-level feature aggregation module is designed to better integrate features across spatial resolutions. To support deployment in real-world scenarios, AutoNAD incorporates a latency-aware prior derived from runtime statistics to guide efficient architecture selection. Extensive experiments on industrial datasets and a real production line confirm the effectiveness and practicality of the proposed framework.
\bibliographystyle{ieeetr}
\bibliography{ref}

\begin{thebibliography}{10}

\bibitem{tmech-autoencoder}
T.~Niu, B.~Li, W.~Li, Y.~Qiu, and S.~Niu, ``Positive-sample-based surface defect detection using memory-augmented adversarial autoencoders,'' {\em IEEE/ASME Trans. Mechatron.}, vol.~27, no.~1, pp.~46--57, 2022.

\bibitem{tmech-multi-view}
J.~Zhou, M.~Liu, Y.~Ma, S.~Jiang, and Y.~Wang, ``Multi-view attention guided feature learning for unsupervised surface defect detection,'' {\em IEEE/ASME Trans. Mechatron.}, pp.~1--9, 2025.

\bibitem{tmech-robust}
X.~Wang, W.~Bian, and X.~Zhao, ``Robust unsupervised anomaly detection for surface defects based on stacked broad learning system,'' {\em IEEE/ASME Trans. Mechatron.}, pp.~1--11, 2024.

\bibitem{tase-RDNet-KD}
W.~Zhou, J.~Yang, W.~Yan, and M.~Fang, ``Rdnet-kd: Recursive encoder, bimodal screening fusion, and knowledge distillation network for rail defect detection,'' {\em IEEE Trans. Autom. Sci. Eng.}, vol.~22, pp.~2031--2040, 2025.

\bibitem{tip-wavelet}
Q.~Zhang, J.~Lai, J.~Zhu, and X.~Xie, ``Wavelet-guided promotion-suppression transformer for surface-defect detection,'' {\em IEEE Trans. Image Process.}, vol.~32, pp.~4517--4528, 2023.

\bibitem{PGA}
H.~Dong, K.~Song, Y.~He, J.~Xu, Y.~Yan, and Q.~Meng, ``Pga-net: Pyramid feature fusion and global context attention network for automated surface defect detection,'' {\em IEEE Trans. Ind. Informat.}, vol.~16, no.~12, pp.~7448--7458, 2020.

\bibitem{tim-boundary}
C.-C. Yeung and K.-M. Lam, ``Attentive boundary-aware fusion for defect semantic segmentation using transformer,'' {\em IEEE Trans. Instrum. Meas.}, vol.~72, pp.~1--13, 2023.

\bibitem{tim-global_local}
L.~Zuo, H.~Xiao, L.~Wen, and L.~Gao, ``A pixel-level segmentation convolutional neural network based on global and local feature fusion for surface defect detection,'' {\em IEEE Trans. Instrum. Meas.}, vol.~72, pp.~1--10, 2023.

\bibitem{tim-Context-Aware}
G.~Zhang, Y.~Lu, X.~Jiang, F.~Yan, and M.~Xu, ``Context-aware adaptive weighted attention network for real-time surface defect segmentation,'' {\em IEEE Trans. Instrum. Meas.}, vol.~73, pp.~1--13, 2024.

\bibitem{tim-mlp}
S.~Yuan, L.~Li, H.~Chen, and X.~Li, ``Surface defect detection of highly reflective leather based on dual-mask-guided deep-learning model,'' {\em IEEE Trans. Instrum. Meas.}, vol.~72, pp.~1--13, 2023.

\bibitem{iccv-cvt}
H.~Wu, B.~Xiao, N.~Codella, M.~Liu, X.~Dai, L.~Yuan, and L.~Zhang, ``Cvt: Introducing convolutions to vision transformers,'' in {\em Proc. IEEE Int. Conf. Comput. Vis. (ICCV)}, pp.~22--31, October 2021.

\bibitem{vits}
A.~Dosovitskiy, L.~Beyer, A.~Kolesnikov, D.~Weissenborn, X.~Zhai, T.~Unterthiner, M.~Dehghani, M.~Minderer, G.~Heigold, S.~Gelly, J.~Uszkoreit, and N.~Houlsby, ``An image is worth 16x16 words: Transformers for image recognition at scale,'' in {\em Proc. Int. Conf. Learn. Represent.}, 2021.

\bibitem{tim-Deep-Feature}
J.~Cao, G.~Yang, and X.~Yang, ``A pixel-level segmentation convolutional neural network based on deep feature fusion for surface defect detection,'' {\em IEEE Trans. Instrum. Meas.}, vol.~70, pp.~1--12, 2021.

\bibitem{tii-RERN}
C.~Wang, H.~Chen, and S.~Zhao, ``Rern: Rich edge features refinement detection network for polycrystalline solar cell defect segmentation,'' {\em IEEE Trans. Ind. Informat.}, vol.~20, no.~2, pp.~1408--1419, 2024.

\bibitem{iccv-Crackformer}
H.~Liu, X.~Miao, C.~Mertz, C.~Xu, and H.~Kong, ``Crackformer: Transformer network for fine-grained crack detection,'' in {\em Proc. IEEE Int. Conf. Comput. Vis. (ICCV)}, pp.~3783--3792, October 2021.

\bibitem{nips-mlp-mixer}
I.~O. Tolstikhin, N.~Houlsby, A.~Kolesnikov, L.~Beyer, X.~Zhai, T.~Unterthiner, J.~Yung, A.~Steiner, D.~Keysers, J.~Uszkoreit, M.~Lucic, and A.~Dosovitskiy, ``Mlp-mixer: An all-mlp architecture for vision,'' in {\em Proc. Adv. Neural Inf. Process. Syst.}, vol.~34, pp.~24261--24272, Curran Associates, Inc., 2021.

\bibitem{rl-1}
B.~Zoph and Q.~V. Le, ``Neural architecture search with reinforcement learning,'' {\em arXiv preprint arXiv:1611.01578}, 2016.

\bibitem{ea-1-icml}
E.~Real, S.~Moore, A.~Selle, S.~Saxena, Y.~L. Suematsu, J.~Tan, Q.~V. Le, and A.~Kurakin, ``Large-scale evolution of image classifiers,'' in {\em Proc. Int. Conf. Mach. Learn.}, vol.~70, pp.~2902--2911, 2017.

\bibitem{one-shot-smash}
A.~Brock, T.~Lim, J.~M. Ritchie, and N.~Weston, ``Smash: one-shot model architecture search through hypernetworks,'' {\em arXiv preprint arXiv:1708.05344}, 2017.

\bibitem{iccv-autoformer}
M.~Chen, H.~Peng, J.~Fu, and H.~Ling, ``Autoformer: Searching transformers for visual recognition,'' in {\em Proc. IEEE Int. Conf. Comput. Vis. (ICCV)}, pp.~12270--12280, October 2021.

\bibitem{eccv-bignas}
J.~Yu, P.~Jin, H.~Liu, G.~Bender, P.-J. Kindermans, M.~Tan, T.~Huang, X.~Song, R.~Pang, and Q.~Le, ``Bignas: Scaling up neural architecture search with big single-stage models,'' in {\em Proc. Eur. Conf. Comput. Vis. (ECCV)}, pp.~702--717, 2020.

\bibitem{cvpr-Auto-DeepLab}
C.~Liu, L.-C. Chen, F.~Schroff, H.~Adam, W.~Hua, A.~L. Yuille, and L.~Fei-Fei, ``Auto-deeplab: Hierarchical neural architecture search for semantic image segmentation,'' in {\em Proc. IEEE Conf. Comput. Vis. Pattern Recognit. (CVPR)}, pp.~82--92, June 2019.

\bibitem{fasterseg}
W.~Chen, X.~Gong, X.~Liu, Q.~Zhang, Y.~Li, and Z.~Wang, ``Fasterseg: Searching for faster real-time semantic segmentation,'' in {\em Proc. Int. Conf. Learn. Represent.}, 2020.

\bibitem{single-one-shot}
Z.~Guo, X.~Zhang, H.~Mu, W.~Heng, Z.~Liu, Y.~Wei, and J.~Sun, ``Single path one-shot neural architecture search with uniform sampling,'' {\em arXiv preprint arXiv:1904.00420}, 2019.

\bibitem{s2mlpv2}
T.~Yu, X.~Li, Y.~Cai, M.~Sun, and P.~Li, ``S$^2$-mlpv2: Improved spatial-shift mlp architecture for vision,'' {\em arXiv preprint arXiv:2108.01072}, 2021.

\bibitem{cvpr-on-the-integra}
X.~Pan, C.~Ge, R.~Lu, S.~Song, G.~Chen, Z.~Huang, and G.~Huang, ``On the integration of self-attention and convolution,'' in {\em Proc. IEEE Conf. Comput. Vis. Pattern Recognit. (CVPR)}, pp.~815--825, June 2022.

\bibitem{tpmai-permutator}
Q.~Hou, Z.~Jiang, L.~Yuan, M.-M. Cheng, S.~Yan, and J.~Feng, ``Vision permutator: A permutable mlp-like architecture for visual recognition,'' {\em IEEE Trans. Pattern Anal. Mach. Intell.}, vol.~45, no.~1, pp.~1328--1334, 2023.

\bibitem{cvpr-fpn}
T.-Y. Lin, P.~Dollar, R.~Girshick, K.~He, B.~Hariharan, and S.~Belongie, ``Feature pyramid networks for object detection,'' in {\em Proc. IEEE Conf. Comput. Vis. Pattern Recognit. (CVPR)}, pp.~2117--2125, July 2017.

\bibitem{eccv-upernet}
T.~Xiao, Y.~Liu, B.~Zhou, Y.~Jiang, and J.~Sun, ``Unified perceptual parsing for scene understanding,'' in {\em Proc. Eur. Conf. Comput. Vis. (ECCV)}, September 2018.

\bibitem{cvpr-psp}
H.~Zhao, J.~Shi, X.~Qi, X.~Wang, and J.~Jia, ``Pyramid scene parsing network,'' in {\em Proc. IEEE Conf. Comput. Vis. Pattern Recognit. (CVPR)}, pp.~2881--2890, July 2017.

\bibitem{SONG2013858}
K.~Song and Y.~Yan, ``A noise robust method based on completed local binary patterns for hot-rolled steel strip surface defects,'' {\em Appl. Surface Sci.}, vol.~285, pp.~858--864, 2013.

\bibitem{FDSNeT}
J.~Zhang, R.~Ding, M.~Ban, and T.~Guo, ``Fdsnet: An accurate real-time surface defect segmentation network,'' in {\em Proc. IEEE Int. Conf. Acoust. Speech Signal Process. (ICASSP)}, pp.~3803--3807, 2022.

\bibitem{Magnetic}
Y.~Huang, C.~Qiu, and K.~Yuan, ``Surface defect saliency of magnetic tile,'' {\em Vis. Comput.}, vol.~36, no.~1, p.~85–96, 2020.

\bibitem{eccv-deeplabv3_plus}
L.-C. Chen, Y.~Zhu, G.~Papandreou, F.~Schroff, and H.~Adam, ``Encoder-decoder with atrous separable convolution for semantic image segmentation,'' in {\em Proc. Eur. Conf. Comput. Vis. (ECCV)}, September 2018.

\bibitem{CVPR-hr_nas}
M.~Ding, X.~Lian, L.~Yang, P.~Wang, X.~Jin, Z.~Lu, and P.~Luo, ``Hr-nas: Searching efficient high-resolution neural architectures with lightweight transformers,'' in {\em Proc. IEEE Conf. Comput. Vis. Pattern Recognit. (CVPR)}, pp.~2982--2992, June 2021.

\bibitem{ICCV-swin}
Z.~Liu, Y.~Lin, Y.~Cao, H.~Hu, Y.~Wei, Z.~Zhang, S.~Lin, and B.~Guo, ``Swin transformer: Hierarchical vision transformer using shifted windows,'' in {\em Proc. IEEE Int. Conf. Comput. Vis. (ICCV)}, pp.~10012--10022, October 2021.

\bibitem{nips-segformer}
E.~Xie, W.~Wang, Z.~Yu, A.~Anandkumar, J.~M. Alvarez, and P.~Luo, ``Segformer: Simple and efficient design for semantic segmentation with transformers,'' in {\em Proc. Adv. Neural Inf. Process. Syst.}, vol.~34, pp.~12077--12090, Curran Associates, Inc., 2021.

\bibitem{cvpr-topformer}
W.~Zhang, Z.~Huang, G.~Luo, T.~Chen, X.~Wang, W.~Liu, G.~Yu, and C.~Shen, ``Topformer: Token pyramid transformer for mobile semantic segmentation,'' in {\em Proc. IEEE Conf. Comput. Vis. Pattern Recognit. (CVPR)}, pp.~12083--12093, June 2022.

\bibitem{ddrnet}
H.~Pan, Y.~Hong, W.~Sun, and Y.~Jia, ``Deep dual-resolution networks for real-time and accurate semantic segmentation of traffic scenes,'' {\em IEEE Trans. Intell. Transp. Syst.}, vol.~24, no.~3, pp.~3448--3460, 2023.

\bibitem{CVPR_pidnet}
J.~Xu, Z.~Xiong, and S.~P. Bhattacharyya, ``Pidnet: A real-time semantic segmentation network inspired by pid controllers,'' in {\em Proc. IEEE Conf. Comput. Vis. Pattern Recognit. (CVPR)}, pp.~19529--19539, June 2023.

\bibitem{tpami-cycle-mlp}
S.~Chen, E.~Xie, C.~Ge, R.~Chen, D.~Liang, and P.~Luo, ``Cyclemlp: A mlp-like architecture for dense visual predictions,'' {\em IEEE Trans. Pattern Anal. Mach. Intell.}, vol.~45, no.~12, pp.~14284--14300, 2023.

\bibitem{SDPT-tits}
H.~Cao, G.~Chen, H.~Zhao, D.~Jiang, X.~Zhang, Q.~Tian, and A.~Knoll, ``Sdpt: Semantic-aware dimension-pooling transformer for image segmentation,'' {\em IEEE Trans. Intell. Transp. Syst.}, vol.~25, no.~11, pp.~15934--15946, 2024.

\bibitem{CVPR-efficient-sam}
Y.~Xiong, B.~Varadarajan, L.~Wu, X.~Xiang, F.~Xiao, C.~Zhu, X.~Dai, D.~Wang, F.~Sun, F.~Iandola, R.~Krishnamoorthi, and V.~Chandra, ``Efficientsam: Leveraged masked image pretraining for efficient segment anything,'' in {\em Proc. IEEE Conf. Comput. Vis. Pattern Recognit. (CVPR)}, pp.~16111--16121, June 2024.

\bibitem{tcsvt-dsnet}
Z.~Guo, L.~Bian, H.~Wei, J.~Li, H.~Ni, and X.~Huang, ``Dsnet: A novel way to use atrous convolutions in semantic segmentation,'' {\em IEEE Trans. Circuits Syst. Video Technol.}, vol.~35, no.~4, pp.~3679--3692, 2025.

\bibitem{ICCV_FairNAS}
X.~Chu, B.~Zhang, and R.~Xu, ``Fairnas: Rethinking evaluation fairness of weight sharing neural architecture search,'' in {\em Proc. IEEE Int. Conf. Comput. Vis. (ICCV)}, pp.~12239--12248, October 2021.

\bibitem{dropout}
N.~Srivastava, G.~Hinton, A.~Krizhevsky, I.~Sutskever, and R.~Salakhutdinov, ``Dropout: A simple way to prevent neural networks from overfitting,'' {\em J. Mach. Learn. Res.}, vol.~15, no.~56, pp.~1929--1958, 2014.

\end{thebibliography}

\end{document}